\documentclass[11pt,a4paper]{article}
\usepackage{theapa, rawfonts}
\usepackage{times}
\usepackage{latexsym}
\usepackage{url}
\usepackage{graphicx}
\usepackage{amsmath}

\begin{document}

   \begin{center}
      \Large\textbf{Language Model Adaptation for Language and Dialect Identification of Text}\\
      \large \it{
\vspace{5 mm}  
    Tommi Jauhiainen \\
  University of Helsinki \\
  tommi.jauhiainen@helsinki.fi \\
\vspace{5 mm}  
  Krister Lind\'en \\   
  University of Helsinki \\
  krister.linden@helsinki.fi \\
\vspace{5 mm}  
  Heidi Jauhiainen \\ 
  University of Helsinki \\
  heidi.jauhiainen@helsinki.fi \\}
   \end{center}

\begin{abstract}
This article describes an unsupervised language model adaptation approach that can be used to enhance the performance of language identification methods. The approach is applied to a current version of the HeLI language identification method, which is now called HeLI 2.0. We describe the HeLI 2.0 method in detail. The resulting system is evaluated using the datasets from the German dialect identification and Indo-Aryan language identification shared tasks of the VarDial workshops 2017 and 2018. The new approach with language identification provides considerably higher F1-scores than the previous HeLI method or the other systems which participated in the shared tasks. The results indicate that unsupervised language model adaptation should be considered as an option in all language identification tasks, especially in those where encountering out-of-domain data is likely.
\end{abstract}

\section{Introduction}

Automatic language identification of text has been researched since the 1960s. It has been considered as a subspecies of general text categorization and most of the methods used are similar to those used in categorizing text according to their topic. However, deep learning techniques have not proven to be as efficient in language identification as they have been in other categorization tasks \cite{medvedeva1}.

For the past six years, we have been developing a language identifying method, which we call HeLI, for the Finno-Ugric Languages and the Internet project \cite{jauhiainen4}. The HeLI method is a supervised general purpose language identification method relying on observations of word and character \emph{n}-gram frequencies from a language labeled corpus. The method is similar to Naive Bayes when using only relative frequencies of words as probabilities. Unlike Naive Bayes, it uses a back-off scheme to approximate the probabilities of individual words if the words themselves are not found in the language models. As language models, we use word unigrams and character level \emph{n}-grams. The optimal combination of the language models used with the back-off scheme depends on the situation and is determined empirically using a development set. The latest evolution of the HeLI method, HeLI 2.0, is described in this article.

One of the remaining difficult cases in language identification is the identification of language varieties or dialects. The task of language identification is less difficult if the set of possible languages does not include very similar languages. If we try to discriminate between very close languages or dialects, the task becomes increasingly more difficult \cite{tiedemann1}. The first ones to experiment with language identification for close languages were Sibun and Reynar \citeA{sibun1} who had Croatian, Serbian, and Slovak as part of their language repertoire. The differences between definitions of dialects and languages are not usually clearly defined, at least not in terms which would be able to help us automatically decide whether we are dealing with languages or dialects. Furthermore, the methods used for dialect identification are most of the time exactly the same as for general language identification. During the last five years, the state-of-the-art language identification methods have been put to the test in a series of shared tasks as part of VarDial workshops \cite{zampieri6,zampieri8,malmasi9,zampieri9,vardial2018report}. We have used the HeLI method and its variations in the shared tasks of the four latest VarDial workshops \cite{jauhiainen-jauhiainen-linden:2015:LT4VarDial,jauhiainen-linden-jauhiainen:2016:VarDial3,jauhiainen-linden-jauhiainen:2017:VarDial,jauhiainen-linden-jauhiainen:2018:DFS,jauhiainen-linden-jauhiainen:2018:GDI,jauhiainen-linden-jauhiainen:2018:ILI}. The HeLI method has proven to be robust and it competes well with other state-of-the-art language identification methods.

Another remaining difficult case in language identification is when the training data is not in the same domain as the data to be identified. Being out-of-domain can mean several things. For example, the training data can be from a different genre, different time period, and/or produced by different writers than the data to be identified. The identification accuracies are considerably lower on out-of-domain data \cite{li3} depending on the degree of out-of-domainness. The extreme example of in-domainness is when the training data and test data are from different parts of the same text as it has been in several language identification experiments in the past \cite{vatanen1,brown1,brown2,brown3}. Classifiers can be more or less sensitive to the domain differences between the training and the testing data depending on the machine learning methods used \cite{blodgett1}. One way to diminish the effects of the phenomena is to create domain-general language models using adversarial supervision which reduces the amount of domain-specific information in the language models \cite{li3}. We suggest that another way is to use active language model adaptation.

In language model (LM) adaptation, we use the unlabelled mystery text itself to enhance the language models used by a language identifier. The language identification method used in combination with the language model adaptation approach presented in this article must be able to produce a confidence score of how well the identification has performed. As the language models are updated regularly while the identification is ongoing, the approach also benefits from the language identification method being non-discriminative. If the method is non-discriminative, all the training material does not have to be re-processed when adding new information into the language models. To our knowledge, language model adaptation has not been used in language identification of digital text before the first versions of the method presented in this article were used in the shared tasks of the 2018 VarDial workshop \cite{jauhiainen-linden-jauhiainen:2018:DFS,jauhiainen-linden-jauhiainen:2018:GDI,jauhiainen-linden-jauhiainen:2018:ILI}. Concurrently with our current work, Ionescu and Butnaru \citeA{ionescu2} presented an adaptive version of the Kernel Ridge Classifier which they evaluated on the Arabic Dialect Identification (ADI) dataset from the 2017 VarDial workshop \cite{zampieri9}.

In this article, we first review the previous work relating to German dialect identification, Indo-Aryan language identification, and language model adaptation (Section~\ref{relatedwork}). We then present the methods used in the article: the HeLI 2.0 method for language identification, three confidence estimation methods, and the algorithm for language model adaptation (Section~\ref{methods}). In Section~\ref{datasets}, we introduce the datasets used for evaluating the methods and, in Section~\ref{experiments}, we evaluate the methods and present the results of the experiments. 

\section{Related work}
\label{relatedwork}

The first automatic language identifier for digital text was described by Mustonen \citeA{mustonen1}. Since this first article, hundreds of conference and journal articles describing language identification experiments and methods have been published. For a recent survey on language identification and the methods used in the literature see Jauhiainen \emph{et al.} \citeA{jauhiainen2018automatic}. The HeLI method was first published in 2010 as part of a master's thesis \cite{jauhiainen1}, and has since been used, outside the VarDial workshops, for language set identification \cite{jauhiainen3} as well as general language identification with a large number of languages \cite{jauhiainen6}. 

\subsection{German dialect identification}

German dialect identification has earlier been considered by Scherrer and Rambow \citeA{scherrer1}, who used a lexicon of dialectal words. Hollenstein and Aepli \citeA{hollenstein1} experimented with a perplexity-based language identifier using character trigrams. They reached an average F-score of 0.66 on sentence level distinguishing between 5 German dialects.

The results of the first shared task on German dialect identification are described by Zampieri \emph{et al.} \citeA{zampieri9}. Ten teams submitted results on the task utilizing a variety of machine learning methods used for language identification. The team \emph{MAZA} \cite{malmasi-zampieri:2017:VarDial1} experimented with different types of support vector machine (SVM) ensembles: plurality voting, mean probability, and meta-classifier. The meta-classifier ensemble using the Random Forest algorithm for classification obtained the best results. The team \emph{CECL} \cite{bestgen:2017:VarDial} used SVMs as well, and their best results were obtained using an additional procedure to equalize the number of sentences assigned to each category. Team \emph{CLUZH} experimented with na{\"i}ve Bayes (NB), conditional random fields (CRF), as well as a majority voting ensemble consisting of NB, CRF, and SVM \cite{clematide-makarov:2017:VarDial}. Their best results were reached using CRF. Team \emph{qcri\_mit} used an ensemble of two SVMs and a stochastic gradient classifier (SGD). Team \emph{unibuckernel} experimented with different kernels using kernel ridge regression (KRR) and kernel discriminant analysis (KDA) \cite{ionescu-butnaru:2017:VarDial}. They obtained their best results using KRR based on the sum of three kernels. Team \emph{tubasfs} \cite{coltekin2} used SVMs with features weighted using sub-linear TF-IDF (product of term frequency and inverse document frequency) scaling. Team \emph{ahaqst} used cross entropy (CE) with character and word \emph{n}-grams \cite{hanani-qaroush-taylor:2017:VarDial}. Team \emph{Citius\_Ixa\_Imaxin} used perplexity with different features \cite{gamallo-pichel-alegria:2017:VarDial}. Team \emph{XAC\_Bayesline} used NB \cite{barbaresi:2017:VarDial} and team \emph{deepCybErNet} Long Short-Term Memory (LSTM) neural networks. We report the F1-scores obtained by the teams in Table~\ref{GDI1list} together with the results presented in this article.

The second shared task on German dialect identification was organized as part of the 2018 VarDial workshop \cite{vardial2018report}. We participated in the shared task with an early version of the method described in this article and our submission using the language model adaptation scheme reached a clear first place \cite{jauhiainen-linden-jauhiainen:2018:GDI}. Seven other teams submitted results on the shared task. Teams \emph{Twist Bytes} \cite{benitesGDI2018}, \emph{T\"ubingen-Oslo} \cite{coltekinGDI2018}, and \emph{GDI\_classification} \cite{ciobanuGDI2018} used SVMs. The team \emph{safina} used convolutional neural networks (CNN) with direct one-hot encoded vectors, with an embedding layer, as well as with a Gated Recurrent Unit (GRU) layer \cite{aliGDI2018}. The team \emph{LaMa} used a voting ensemble of eight classifiers. The best results for the team \emph{XAC} were achieved using Na\"ive Bayes, but they experimented with Ridge regression and SGD classifiers as well \cite{barbaresiGDI2018}. The team \emph{dkosmajac} used normalized Euclidean distance. After the shared task, the team \emph{Twist Bytes} was able to slightly improve their F1-score by using a higher number of features \cite{benitesGDI2018}. However, the exact number of included features was not determined using the development set, but it was the optimal number for the test set. Using the full set of features resulted again in a lower score. We report the F1-scores obtained by the teams in Table~\ref{GDI2list} together with the results obtained in this article.

\subsection{Language identification for Devanagari script}

Language identification research in distinguishing between languages using the Devanagari script is much more uncommon than for the Latin script. However, there has been some research already before the Indo-Aryan Language Identification (ILI) shared task at VarDial 2018 \cite{vardial2018report}. Kruengkrai \emph{et al.} \citeA{kruengkrai1} presented results from language identification experiments between ten Indian languages, including four languages written in Devanagari: Sanskrit, Marathi, Magahi, and Hindi. For the ten Indian languages they obtained over 90\% accuracy with 70-byte long mystery text sequences. As language identification method, they used SVMs with string kernels. Murthy and Kumar \citeA{murthy1} compared the use of language models based on bytes with models based on aksharas. Aksharas are the syllables or orthographic units of the Brahmi scripts \cite{vaid1}. After evaluating the language identification between different pairs of languages, they concluded that the akshara-based models perform better than byte-based. They used multiple linear regression as the classification method.

Sreejith \emph{et al.} \citeA{sreejith1} tested language identification with Markovian character and word \emph{n}-grams from one to three with Hindi and Sanskrit. A character bigram-based language identifier fared the best and managed to gain an accuracy of 99.75\%  for sentence-sized mystery texts. Indhuja \emph{et al.} \citeA{indhuja2} continued the work of Sreejith \emph{et al.} \citeA{sreejith1} investigating the language identification between Hindi, Sanskrit, Marathi, Nepali, and Bhojpuri. In a similar fashion, they evaluated the use of Markovian character and word \emph{n}-grams from one to three. For this set of languages, word unigrams performed the best, obtaining 88\% accuracy with the sentence-sized mystery texts.

Bergsma \emph{et al.} \citeA{bergsma1} collected tweets in three languages written with the Devanagari script: Hindi, Marathi, and Nepali. They managed to identify the language of the tweets with 96.2\% accuracy using a logistic regression (LR) classifier \cite{hosmer1} with up to 4-grams of characters. Using an additional training corpus, they reached 97.9\% accuracy with the A-variant of prediction by partial matching (PPM). Later, Pla and Hurtado \citeA{pla2} experimented with the corpus of Bergsma \emph{et al.} \citeA{bergsma1}. Their approach using words weighted with TF-IDF and SVMs reached 97.7\% accuracy on the tweets when using only the provided tweet training corpus. Hasimu and Silamu \citeA{hasimu3} included the same three languages in their test setting. They used a two-stage language identification system where the languages were first identified as a group using Unicode code ranges. In the second stage, the languages written with the Devanagari script were individually identified using SVMs with character bigrams. Their tests resulted in an F1-score of 0.993 within the group of languages using Devanagari with 700 best distinguishing bigrams. Indhuja \emph{et al.} \citeA{indhuja2} provided test results for several different combinations of the five languages and for the set of languages used by Hasimu and Silamu \citeA{hasimu3}, they reached 96\% accuracy with word unigrams.

Rani \emph{et al.} \citeA{RANI18.16} described a language identification system which they used for discriminating between Hindi and Magahi. Their language identifier using lexicons and suffixes of three characters obtained an accuracy of 86.34\%. Kumar \emph{et al.} \citeA{KUMAR18.26} provided an overview of experiments on an earlier version of the dataset used in the ILI shared task including five closely related Indo-Aryan languages: Awadhi, Bhojpuri, Braj, Hindi, and Magahi. They managed to obtain an accuracy of 96.48\% and a macro F1-score of 0.96 on the sentence level. For sentence level language identification, these results are quite good and as such they indicate that the languages, at least in their written form as evidenced by the corpus, are not as closely related as for example the Balkan languages: Croatian, Serbian, and Bosnian.

The results of the first shared task on Indo-Aryan language identification are described by Zampieri \emph{et al.} \citeA{vardial2018report}. Eight teams submitted results on the task. As for the 2nd edition of the GDI shared task, we participated with an early version of the method described in this article. Again, our submission using a language model adaptation scheme reached a clear first place \cite{jauhiainen-linden-jauhiainen:2018:ILI}. Seven other teams submitted results on the shared task. The team with the second best results, \emph{T\"ubingen-Oslo}, submitted their best results using SVMs \cite{coltekinGDI2018}. In addition to the SVMs, they experimented with Recurrent Neural Networks (RNN) with GRUs and LSTMs but their RNNs never achieved results comparable to the SVMs. The team \emph{ILIdentification} used an SVM ensemble \cite{ciobanuILI2018}. 
The best results for the team \emph{XAC} were achieved using Ridge regression \cite{barbaresiGDI2018}. In addition to Ridge regression, they experimented with NB and SGD classifiers, which did not perform as well. 
The team \emph{safina} used CNNs with direct one-hot encoded vectors, with an embedding layer, as well as with a GRU layer \cite{aliILI2018}. The team \emph{dkosmajac} used normalized Euclidean distance. The team \emph{we\_are\_indian} used word-level LSTM RNNs in their best submission and statistical \emph{n}-gram approach with mutual information in their second submission \cite{guptaILI2018}. The team \emph{LaMa} used NB. We report the F1-scores obtained by the teams in Table~\ref{ILI1list} together with the results presented in this article.

\subsection{Language model adaptation}

Even though language model adaptation has not been used in language identification of text in the past, it has been used in other areas of natural language processing. Jelinek \emph{et al.} \citeA{jelinek1} used a dynamic LM and Bacchiani and Roark \citeA{bacchiani1} used self-adaptation on a test set in speech recognition. Bacchiani and Roark \citeA{bacchiani1} experimented with iterative adaptation on their language models and noticed that one iteration made the results better but that subsequent iterations made them worse. Zlatkova \emph{et al.} \citeA{zlatkova1} used a Logistic Regression classifier in the Style Change Detection shared task \cite{kestemont1}. Their winning system  fitted their TF-IDF features on the testing data in addition to the training data.

Language model adaptation was used by Chen and Liu \citeA{chen1} for identifying the language of speech. In the system built by them, the speech is first run through Hidden Markov Model-based phone recognizers (one for each language) which tokenize the speech into sequences of phones. The probabilities of those sequences are calculated using corresponding language models and the most probable language is selected. An adaptation routine is then used so that each of the phonetic transcriptions of the individual speech utterances is used to calculate probabilities for words \(t\), given a word \emph{n}-gram history of \(h\) as in Equation~\ref{chent}.

\begin{equation}
P_{a}(t|h) = \lambda P_{o}(t|h)+(1-\lambda)P_{n}(t|h) \text{,}
\label{chent}
\end{equation}

\vspace{2mm}

\noindent where \(P_{o}\) is the original probability calculated from the training material, \(P_{n}\) the probability calculated from the data being identified, and \(P_{a}\) the new adapted probability. \(\lambda\) is the weight given to original probabilities. This adaptation method resulted in decreasing the error rate in three-way identification between Chinese, English, and Russian by 2.88\% and 3.84\% on an out-of-domain (different channels) data and by 0.44\% on in-domain (same channel) data.

Later, also Zhong \emph{et al.} \citeA{zhong1} used language model adaptation in language identification of speech. They evaluated three different confidence measures and the best faring measure \(C\) is defined as follows:

\begin{equation}
C(g_i,M) = \frac{1}{n}[\log(P(M|g_i))-\log(P(M|g_j))] \text{,}
\label{zhong1}
\end{equation}

\vspace{2mm}

\noindent where \(M\) is the sequence to be identified, \(n\) the number of frames in the utterance, \(g_i\) the best identified language, and \(g_j\) the second best identified language. The two other evaluated confidence measures were clearly inferior. Although the \(C(g_i,M)\) measure performed the best of the individual measures, a Bayesian classifier-based ensemble using all the three measures gave slightly higher results. Zhong \emph{et al.} \citeA{zhong1} use the same language adaptation method as Chen and Liu \citeA{chen1}, using the confidence measures to set the \(\lambda\) for each utterance.

We used an early version of the language model adaptation technique presented in this article in three of the 2018 VarDial workshop shared tasks \cite{jauhiainen-linden-jauhiainen:2018:DFS,jauhiainen-linden-jauhiainen:2018:GDI,jauhiainen-linden-jauhiainen:2018:ILI}.

The adaptive language identification method presented by Ionescu and Butnaru \citeA{ionescu2} improved the accuracy from 76.27\% to 78.35\% on the ADI dataset. In their method, they retrain the language models once by adding 1,000 of the best identified (sorted by the confidence scores produced by their language identification method) unlabelled test samples to the training data.

\section{The Methods}
\label{methods}

In this section, we present the detailed descriptions of the methods used in the experiments. First we describe HeLI 2.0, the language identification method used. Then we present the confidence measures we consider in this article. We conclude this section by describing the language model adaptation method used.

\subsection{Language Identification}
\label{sec:Proposed}

We use the HeLI method \cite{jauhiainen-linden-jauhiainen:2016:VarDial3} for language identification. The HeLI method has been rivalling SVMs already before the language model adaptation was added, reaching a shared first place in the 2016 Discriminating Similar Languages (DSL) shared task \cite{malmasi9}. The HeLI method is mostly non-discriminative and it is relatively quick to incorporate new material into the LMs of the language identifier. We have made a modification to the method where the original penalty value is replaced with a smoothing value that is calculated from the sizes of the LMs. This modification is needed especially for such cases where the language models grow considerably because of LM adaptation, as the original penalty value was depending on the sizes of the training corpus during the development phase. The penalty modifier \(p_{mod}\) is introduced to penalize those languages where features encountered during the identification are absent. The \(p_{mod}\) parameter is optimized using the development corpus and in the experiments presented in this article, the optimal value varies between 1.09 and 1.16. The complete formula for the HeLI 2.0 method is presented here and we provide the modified equations for the values used in the LMs in a similar notation as that used by Jauhiainen \emph{et al.} \citeA{jauhiainen-linden-jauhiainen:2016:VarDial3}.

The method aims to determine the language \(g \in G\) in which the mystery text \(M\) has been written, when all languages in the set \(G\) are known to the language identifier. Each language is represented by several different language models only one of which is used for every word \(t\) found in the mystery text. The language models for each language are: a model\footnote{There can be several models for words, depending on the preprocessing scheme.} based on words and one or more models based on character \emph{n}-grams from one to \(n_{max}\). The mystery text is processed one word at a time. The word-based models are used first and if an unknown word is encountered in the mystery text, the method backs off to using the character \emph{n}-grams of the size \(n_{max}\). If it is not possible to apply the character \emph{n}-grams of the size \(n_{max}\), the method backs off to lower order character \emph{n}-grams and, if needed, continues backing off until character unigrams.

{\bf Creating the language models}: The training data is preprocessed in different ways to produce different types of language models. The most usual way is to lowercase the text and tokenize it into words using non-alphabetic and non-ideographic characters as delimiters. It is possible to generate several language models for words using different preprocessing schemes, and then use the development material to determine which models and in which back-off order are usable for the current task. 

The relative frequencies of the words are calculated. Also the relative frequencies of character \emph{n}-grams from 1 to \(n_{max}\) are calculated inside the words, so that the preceding and the following space-characters are included\footnote{A space character is added to the beginning and the end of each word even if it was not there originally.}. The character \emph{n}-grams are overlapping, so that for example a word with three characters includes three character trigrams. Word \emph{n}-grams were not used in the experiments of this article, so all subsequent references to \emph{n}-grams in this article refer to \emph{n}-grams of characters. After calculating the relative frequencies, we transform those relative frequencies into scores using 10-based logarithms.

The corpus containing only the word tokens in the language models is called \(C\). A corpus \(C\) in language \(g\) is denoted by \(C_{g}\). \(dom(O(C))\) is the set of all words found in the models of any of the languages \(g \in G\). For each word \(t \in dom(O(C))\), the values \(v_{C_{g}}(t)\) for each language \(g\) are calculated, as in Equation~\ref{neglog10token}.

\begin{equation}
v_{C_{g}}(t) = \left\{ 
 \begin{array}{l l}
    -\log_{10}\Big(\frac{c(C_{g},t)}{l_{C_{g}}}\Big) & \text{, if }c(C_{g},t)>0\\
    -\log_{10}\Big(\frac{1}{l_{C_{g}}}\Big)p_{mod} & \text{, if }c(C_{g},t)=0
  \end{array} \right.
\label{neglog10token}
\end{equation}

\vspace{2mm}

\noindent where \(c(C_{g},t)\) is the number of words \(t\) and \(l_{C_{g}}\) is the total number of all words in language \(g\). The parameter \(p_{mod}\) is the penalty modifier which is determined empirically using the development set.

The corpus containing the \emph{n}-grams of the size \(n\) in the language models is called \(C^{n}\). The domain \(dom(O(C^{n}))\) is the set of all character \emph{n}-grams of length \(n\) found in the models of any of the languages \(g \in G\). The values \(v_{C_{g}^{n}}(u)\) are calculated in the same way for all \emph{n}-grams \(u \in dom(O(C^{n}))\) for each language \(g\), as shown in Equation~\ref{neglog10}.

\begin{equation}
v_{C_{g}^{n}}(u) = \left\{ 
  \begin{array}{l l}
    -\log_{10}\Big(\frac{c(C_{g}^{n},u)}{l_{C_{g}^{n}}}\Big) & \text{, if }c(C_{g}^{n},u)>0\\
    -\log_{10}\Big(\frac{1}{l_{C_{g}^{n}}}\Big)p_{mod} & \text{, if }c(C_{g}^{n},u)=0
  \end{array} \right.
\label{neglog10}
\end{equation}

\vspace{2mm}

\noindent where \(c(C_{g}^{n},u)\) is the number of \emph{n}-grams \(u\) found in the corpus of the language \(g\) and \(l_{C_{g}^{n}}\) is the total number of the \emph{n}-grams of length \(n\) in the corpus of language \(g\). These values are used when scoring the words while identifying the language of a text.

{\bf Scoring the text:} The mystery text \(M\) is tokenized into words using the same tokenization scheme as when creating the language models. The words are lowercased when lowercased models are being used. After this, a score \(v_{g}(t)\) is calculated for each word \(t\) in the mystery text for each language \(g\). If the word \(t\) is found in the set of words \(dom(O(C_{g}))\), the corresponding value \(v_{C_{g}}(t)\) for each language \(g\) is assigned as the score \(v_{g}(t)\), as shown in Equation~\ref{assigntokenscore}.

\begin{equation}
\label{assigntokenscore}
v_{g}(t)= \left\{ 
  \begin{array}{l l}
v_{C_{g}}(t) & \quad \text{, if } t \in dom(O(C_{g}))\\
v_{g}(t,min(n_{max},l_t+2)) & \quad \text{, if } t \notin dom(O(C_{g}))
  \end{array} \right.
\end{equation}

\vspace{2mm}

If a word \(t\) is not found in the set of words \(dom(O(C_{g}))\) and the length of the word \(l_t\) is at least \(n_{max}-2\), the language identifier backs off to using character \emph{n}-grams of the length \(n_{max}\). In case the word \(t\) is shorter than \(n_{max}-2\) characters, \(n=l_t+2\).

When using \emph{n}-grams, the word \(t\) is split into overlapping \emph{n}-grams of characters \(u^{n}_{i}\text{, where }i=1, ..., l_t-n\), of the length \(n\). Each of the \emph{n}-grams \(u^{n}_{i}\) is then scored separately for each language \(g\) in the same way as the words.

If the \emph{n}-gram \(u^{n}_{i}\) is found in \(dom(O(C^{n}_g))\), the values in the models are used. If the \emph{n}-gram \(u^{n}_{i}\) is not found in any of the models, it is simply discarded. We define the function \(d_{g}(t,n)\) for counting \emph{n}-grams in \(t\) found in a model in Equation~\ref{eq:dltn}.

\begin{equation}
\label{eq:dltn}
d_{g}(t,n) = \sum^{l_t-n}_{i=1} \left\{
\begin{array}{l l}
1 & \quad \text{, if }  u^n_i \in dom(O(C^n)) \\
0 & \quad \text{, otherwise}
\end{array} \right.
\end{equation}

\vspace{2mm}

When all the \emph{n}-grams of the size \(n\) in the word \(t\) have been processed, the word gets the value of the average of the scored \emph{n}-grams \(u^{n}_{i}\) for each language, as in Equation~\ref{assigntokenscore2}.

\begin{equation}
\label{assigntokenscore2}
v_{g}(t,n) = \left\{
\begin{array}{l l}
\frac{1}{d_{g}(t,n)}\sum_{i=1}^{l_t-n} v_{C^{n}_{g}}(u^{n}_{i}) & \quad \text{, if }  d_{g}(t,n)>0 \\
v_{g}(t,n-1) & \quad \text{, otherwise,}
\end{array} \right.
\end{equation}

\vspace{2mm}

\noindent where \(d_{g}(t,n)\) is the number of \emph{n}-grams \(u^{n}_{i}\) found in the domain \(dom(O(C^{n}_{g}))\). If all of the \emph{n}-grams of the size \(n\) were discarded, \(d_{g}(t,n) = 0\), the language identifier backs off to using \emph{n}-grams of the size \(n-1\).

The whole mystery text \(M\) gets the score \(R_{g}(M)\) equal to the average of the scores of the words \(v_{g}(t)\) for each language \(g\), as in Equation~\ref{mysteryscore}.

\begin{equation}
\label{mysteryscore}
R_{g}(M)=\frac{\sum_{i=1}^{l_{T(M)}} v_{g}(t_{i}) }{l_{T(M)}}
\end{equation}

\vspace{2mm}

\noindent where \(T(M)\) is the sequence of words and \(l_{T(M)}\) is the number of words in the mystery text \(M\). Since we are using negative logarithms of probabilities, the language having the lowest score is returned as the language with the maximum probability for the mystery text.

\subsection{Confidence estimation}
\label{confidenceequations}

In order to be able to select the best candidate for language model adaptation, the language identifier needs to provide a confidence score for the identified language. We evaluated three different confidence measures that seemed applicable to the HeLI 2.0 method. 

In the first measure, we estimate the confidence of the identification as the difference between the scores \(R(M)\) of the best and the second best identified language. Zhong \emph{et al.} \citeA{zhong1} call this confidence score \(CM_{BS}\) and in our case it is calculated using the following equation:

\begin{equation}
CM_{BS}(M) = R_{h}(M) - R_{g}(M) 
\label{cmbs}
\end{equation}

\vspace{2mm}

\noindent where \(g\) is the best scoring language and \(h\) the second best scoring language.

The second confidence measure, \(CM_{AVG}\), was presented by Chen and Liu \citeA{chen1}. \(CM_{AVG}\) adapted to our situation is calculated as follows:

\begin{equation}
CM_{AVG}(M) = \frac{1}{l_G-1} \sum^{l_G}_{j=1,j\neq g}R_{j}(M) - R_{g}(M) 
\label{chenconfidence}
\end{equation}

\vspace{2mm}

\noindent The third measure, \(CM_{POST}\), presented by Zhong \emph{et al.} \citeA{zhong1}, is calculated with the following equation:

\begin{equation}
CM_{POST}(M) = \log  \sum^{l_G}_{j=1} e^{R_{j}(M)} - R_{g}(M) 
\label{postconfidence}
\end{equation}

\vspace{2mm}

\subsection{Language model adaptation algorithm}

In the first step of our adaptation algorithm, all the mystery texts \(M\) in the mystery text collection \(MC\) (for example, a test set) are preliminarily identified using the HeLI 2.0 method. They are subsequently ranked by their confidence scores \(CM\) and the preliminarily identified collection is split into \(k-q\) parts \(MC_{1 ... k}\). \(k\) is a number between 1 and the total number of mystery texts, \(l_{MC}\), depending on in how many parts we want to split the mystery text collection.\footnote{The only difference between the language adaptation method presented here and the earlier version of the method we used at the shared tasks is that in the shared tasks, the \(k\) was always equal to \(l_{MC}\).} The higher \(k\) is, the longer the identification of the whole collection will take. The number of finally identified parts is \(q\), which in the beginning is \(0\). After ranking, the part \(MC_{1}\) includes the most confidently identified texts and \(MC_{k-q}\) the least confidently identified texts.

Words and character \emph{n}-grams up to the length \(n_{max}\) are extracted from each mystery text in \(MC_{1}\) and added to the respective language models. Then, all the mystery texts in the part \(MC_{1}\) are set as finally identified and \(q\) is increased by 1. 

Then for as long as \(q<k\), the process is repeated using the newly adapted language models to perform a new preliminary identification for those texts that are not yet finally identified. In the end, all features from all of the mystery texts are included in the language model. This constitutes one epoch of adaptation.

In iterative language model adaptation, the previous algorithm is repeated from the beginning several times.

\section{Test setting}
\label{datasets}

We evaluate the methods presented in the previous section using three standard datasets. The first two datasets are from the GDI shared tasks held at VarDials 2017 and 2018. The third dataset is from the ILI shared task held at VarDial 2018.

\subsection{GDI 2017 dataset}
\label{GDI2017dataset}

The dataset used in the GDI 2017 shared task consists of manual transcriptions of speech utterances by speakers from different areas in Switzerland: Bern, Basel, Lucerne, and Zurich. The variety of German spoken in Switzerland is considered to be a separate language (Swiss German, \emph{gsw}) by the ISO-639-3 standard \cite{lewis1} and these four areas correspond to separate varieties of it. The transcriptions in the dataset are written entirely in lowercased letters. Samard\v{z}i{\'c} {\it et al.} \citeA{samardzic2016archimob} describe the ArchiMob corpus, which is the source for the shared task dataset. Zampieri \emph{et al.} \citeA{zampieri9} describe how the training and test sets were extracted from the ArchiMob corpus for the 2017 shared task. The sizes of the training and test sets can be seen in Table~\ref{setsizesGDI2017}. The shared task was a four-way language identification task between the four German dialects present in the training set.

\begin{table}[h]
\small
\centering
\begin{tabular}{lcc}
\hline
\bf Variety (code) & \bf Training & \bf Test\\
\hline
Bern (BE) & 28,558 & 7,025 \\
Basel (BS) & 28,680 & 7,064 \\
Lucerne (LU) & 28,653 & 7,509\\
Zurich (ZH) & 28,715 & 7,949\\
\hline
\end{tabular}
\caption{List of the Swiss German varieties used in the datasets distributed for the 2017 GDI shared task. The sizes of the training and the test sets are in words.}
\label{setsizesGDI2017}
\end{table}

\subsection{GDI 2018 dataset}
\label{GDI2018dataset}

The dataset used in the GDI 2018 shared task was similar to the one used in GDI 2017. The sizes of the training, the development, and the test sets can be seen in Table~\ref{setsizesGDI2018}. The first track of the shared task was a standard four-way language identification between the four German dialects present in the training set. The GDI 2018 shared task included an additional second track dedicated to unknown dialect detection. The unknown dialect was not included in the training nor the development sets, but it was present in the test set. The test set was identical for both tracks, but the lines containing an unknown dialect were ignored when calculating the results for the first track.

\begin{table}[h]
\small
\centering
\begin{tabular}{lccc}
\hline
\bf Variety (code) & \bf Training & \bf Development & \bf Test\\
\hline
Bern (BE) & 28,558 & 7,404 & 12,013\\
Basel (BS) & 27,421 & 9,544 & 9,802\\
Lucerne (LU) & 29,441 & 8,887 & 11,372\\
Zurich (ZH) & 28,820 & 8,099 & 9,610\\
Unknown dialect (XY) & & & 8,938\\
\hline
\end{tabular}
\caption{List of the Swiss German varieties used in the datasets distributed for the 2018 GDI shared task. The sizes of the training, the development, and the test sets are in words.}
\label{setsizesGDI2018}
\end{table}

\subsection{ILI 2018 dataset}
\label{ILI2018dataset}

The dataset used for the ILI 2018 shared task included text in five languages: Bhojpuri, Hindi, Awadhi, Magahi, and Braj. As can be seen in Table~\ref{ILIlanguages}, there was considerably less training material for the Awadhi language than the other languages. The training corpus for Awadhi had only slightly over 9,000 lines whereas the other languages had around 15,000 lines of text for training. An early version of the dataset, as well as its creation, was described by Kumar \emph{et al.} \citeA{KUMAR18.26}. The ILI 2018 shared task was an open one, allowing the use of any additional data or means. However, we have not used any external data and our results would be exactly the same on a closed version of the task.

\begin{table}[h]
\small
\centering
\begin{tabular}{lccc}
\hline
\bf Language (code) & \bf Training  & \bf Development & \bf Test\\
\hline
Bhojpuri (BHO) & 258,501 & 56,070 & 50,206\\
Hindi (HIN) & 325,458 & 44,215 & 35,863\\
Awadhi (AWA) & 123,737 & 19,616 & 22,984 \\
Magahi (MAG) & 234,649 & 37,809 & 35,358\\
Braj (BRA) & 249,243 & 40,023 & 31,934\\
\hline
\end{tabular}
\caption{List of the Indo-Aryan languages used in the datasets distributed for the 2018 ILI shared task. The sizes of the training, the development, and the test sets are in words.}
\label{ILIlanguages}
\end{table}

\section{Experiments and results}
\label{experiments}

In our experiments, we evaluate the HeLI 2.0 method, the HeLI 2.0 method using language model adaptation, as well as the iterative version of the adaptation. We test all three methods with all of the datasets described in the previous section. First we evaluate the confidence measures using the GDI 2017 dataset and afterwards we use the best performing confidence measure in all further experiments.

We are measuring language identification performance using the macro and the weighted F1-scores. These are the same performance measures that were used in the GDI 2017, GDI 2018, and ILI 2018 shared tasks \cite{zampieri9,vardial2018report}. F1-score is calculated using the precision and the recall as in Equation~\ref{fscore}.

\begin{equation}
\label{fscore}
\text{F1-score}=2*\frac{\text{precision} * \text{recall}}{\text{precision} + \text{recall}}
\end{equation}

\vspace{2mm}

The macro F1-score is the average of the individual F1-scores for the languages and the weighted F1-score is similar, but weighted by the number of instances for each language.

\subsection{Evaluating the confidence measures}

\noindent We evaluated the three confidence measures presented in Section~\ref{confidenceequations} using the GDI 2017 training data. The results of the evaluation are presented in Table~\ref{tab:Confidences}. The underlying data for the table consists of pairs of confidence values and the corresponding Boolean values indicating whether the identification results were correct or not. The data has been ordered according to their confidence score for each of the three measures. The first column in the table tells the percentage of examined top scores. The other columns give the average accuracy in that examined portion of identification results for each confidence measure.

The first row tells us that in the 10\% of the highest confidence identification results according to the \(CM_{BS}\)-measure, 98.5\% of the performed identifications were correct. The two other confidence measures on the other hand fail to arrange the identification results so that the most confident 10\% would be the most accurate 10\%. As a whole, this experiment tells us that the \(CM_{BS}\)-measure is stable and performs well when compared with the other two.

\begin{table}[h]
\small
\center
\begin{tabular}{lccc}
\hline
\bf \% most & \bf \(\mathbf{CM_{BS}}\) & \bf \(\mathbf{CM_{AVG}}\) & \bf \(\mathbf{CM_{POST}}\) \\
\bf confident & \bf accuracy & \bf accuracy & \bf accuracy \\
\hline
0-10\% & 98.5\% & 95.6\% & 88.3\%\\
10-20\% & 98.3\% & 96.3\% & 91.4\%\\
20-30\% & 98.2\% & 96.2\% & 92.6\%\\
30-40\% & 98.2\% & 95.1\% & 92.1\%\\
40-50\% & 97.8\% & 94.6\% & 92.0\%\\
50-60\% & 96.9\% & 94.9\% & 91.7\%\\
60-70\% & 96.0\% & 94.0\% & 91.5\%\\
70-80\% & 94.4\% & 93.2\% & 91.0\%\\
80-90\% & 92.4\% & 91.7\% & 89.9\%\\
90-100\% & 89.0\% & 89.0\% & 89.0\%\\
\hline
\end{tabular}
\caption{Average accuracies within the 10\% portions when the results are sorted by the confidence scores \(CM\).}
\label{tab:Confidences}
\end{table}

In addition to evaluating each individual confidence measure, Zhong \emph{et al.} \citeA{zhong1} evaluated an ensemble combining all of the three measures, gaining somewhat better results than with the otherwise best performing \(CM_{BS}\)-measure. However, in their experiments the two other measures were much more stable than in ours. We decided to use only the simple and well performing \(CM_{BS}\)-measure with our LM adaptation algorithm in the following experiments.

\subsection{Experiments on the GDI 2017 dataset}

\subsubsection{Baseline results and parameter estimation}

As there was no separate development set provided for the GDI 2017 shared task, we divided the training set into training and development partitions. The last 500 lines from the original training data for each language was used for development. The development partition was then used to find the best parameters for the HeLI 2.0 method using the macro F1-score as the performance measure. The macro F1-score is equal to the weighted F1-score, which was used as a ranking measure in the shared task, when the number of tested instances in each class are equal. On the development set, the best macro F1-score of 0.890 was reached with the language identifier where \(n_{max}=5\) words being used and \(p_{mod}=1.16\). We then used the whole training set to train the LMs. On the test set, the language identifier using the same parameters reached the macro F1-score of 0.659, and the weighted F1-score of 0.639.

\subsubsection{Experiments with language model adaptation}

First, we determined the best value for the number of splits \(k\) using the development partition. Table~\ref{tab:koos} shows the increment of the weighted F1-score with different values of \(k\) on the development partition using the same parameters with the HeLI 2.0 method as for the baseline. The results with \(k=1\) are always equal to the baseline. If \(k\) is very high the identification becomes computationally costly as the number of identifications grows exponentially in proportion to \(k\). The absolute increase of the F1-score on the development partition was 0.01 when using \(k=45\).

\begin{table}[h]
\small
\center
\begin{tabular}{lc|lc}
\hline
\bf \emph{k} & \bf weighted F1-score & \bf \emph{k} & \bf weighted F1-score\\
\hline
1 or 2 & 0.890 & 30, 35, or 40 & 0.899\\
3 & 0.893 & 45 & 0.900\\
4 & 0.894 & 50 & 0.899\\
5 & 0.896 & 60 & 0.900\\
10 or 15 & 0.899 & 70, 80, 90, \\
20 or 25 & 0.898 & 100, 150, or 200 & 0.899\\
\hline
\end{tabular}
\caption{The weighted F1-scores obtained by the identifier using LM adaptation with different values of \(k\) when tested on the development partition of the GDI 2017 dataset.}
\label{tab:koos}
\end{table}

\subsubsection{Experiments with thresholding}

We experimented with setting a confidence threshold for the inclusion of new data into the language models. Table~\ref{tab:cfthreshold} shows the results on the development partition. The results show, that there is no confidence score that could be used for thresholding, at least not with the development partition of GDI 2017.

\begin{table}[h]
\small
\center
\begin{tabular}{lc|lc}
\hline
\bf Conf. threshold & \bf macro F1-score & \bf Conf. threshold & \bf macro F1-score\\
\hline
0.00 - 0.01 & 0.900 & 0.10 & 0.898\\
0.02 & 0.899 & 0.16 & 0.896\\
0.04 & 0.898 & 0.32 & 0.889\\
0.06 or 0.08 & 0.899\\
\hline
\end{tabular}
\caption{Weighted F1-scores with confidence threshold for LM adaptation on the development set.}
\label{tab:cfthreshold}
\end{table}

\subsubsection{Results of the LM adaptation on the test data}

Based on the evaluations using the development partition, we decided to use \(k=45\) for the test run. All the training data was used for the initial LM creation. The language identifier using LM adaptation reached the macro F1-score of 0.689 and the weighted F1-score of 0.687 on the test set. The weighted F1-score was 0.048 higher than the one obtained by the non-adaptive version and clearly higher than the other results obtained using the GDI 2017 dataset.

\subsubsection{Iterative adaptation}

We tested repeating the LM adaptation algorithm for several epochs and the results of those trials on the development partition can be seen in Table~\ref{tab:epochs}. The improvement of 0.003 on the original macro F1-score using 13-956 epochs was still considerable. The results seem to indicate that the language models become very stable with repeated adaptation.

\begin{table}[h]
\small
\center
\begin{tabular}{lc|lc}
\hline
\bf Number of epochs & \bf Macro F1-score & \bf Number of epochs & \bf Macro F1-score\\
\hline
1 & 0.900 & 13-956 & 0.903\\
2-4 & 0.901 & 957-999 & 0.902\\
5-12 & 0.902\\
\hline
\end{tabular}
\caption{Macro F1-scores with iterative LM adaptation on the development partition.}
\label{tab:epochs}
\end{table}

We decided to try iterative LM adaptation using 485 epochs with the test set. The tests resulted in a weighted F1-score of 0.700, which was a further 0.013 increase on top of the score obtained without additional iterations. We report the weighted F1-scores from the GDI 2017 shared task together with our own results in Table~\ref{GDI1list}. The methods used are listed in the first column, used features in the second column, and the best reached weighted F1-score in the third column. The results from this paper are bolded. The results using other methods (team names are in parentheses) are collected from the shared task report \cite{zampieri9} as well as from the individual system description articles. The 0.013 point increase obtained with the iterative LM adaptation over the non-iterative version might seem small when compared with the overall increase over the scores of the HeLI 2.0 method, but the increase is still more than the difference between the 1st and 3rd best submitted methods on the original shared task.

\begin{table}[h]
\small
\center
\begin{tabular}{lcc}
\hline
\bf Method (Team) & \bf Features used & \bf wgh. F1\\
\hline
\bf HeLI 2.0 + iterative LM-adapt. & ch. \emph{n}-grams 1-5 and words & \bf  0.700\\
\bf HeLI 2.0 + LM-adapt. & ch. \emph{n}-grams 1-5 and words & \bf  0.687\\
SVM meta-classifier ensemble (MAZA)& ch. \emph{n}-grams 1-6 and words & 0.662\\
SVM, cat. equal. 2 (CECL)&  BM25 ch. \emph{n}-grams 1-5 & 0.661\\
CRF (CLUZH)& ch. \emph{n}-grams, affixes... & 0.653 \\
NB, CRF, and SVM ensemble (CLUZH)& ch. \emph{n}-grams, affixes... & 0.653 \\
SVM probability ensemble (MAZA)& ch. \emph{n}-grams 1-6 and words & 0.647\\
SVM + SGD ensemble (qcri\_mit)& \emph{n}-grams 1-8 & 0.639\\
\bf HeLI 2.0 & ch. \emph{n}-grams 1-5 and words & \bf 0.639\\
SVM, cat. equal. 1 (CECL)&  BM25 ch. \emph{n}-grams 1-5 & 0.638\\
KRR, sum of 3 kernels (unibuckernel)& \emph{n}-grams 3-6 & 0.637\\ 
KDA, sum of 2 kernels (unibuckernel)& \emph{n}-grams 3-6 & 0.635\\ 
KDA, sum of 3 kernels (unibuckernel)& \emph{n}-grams 3-6 & 0.634\\ 
SVM voting ensemble (MAZA)& ch. \emph{n}-grams 1-6 and words & 0.627\\
Linear SVM (tubasfs) & TF-IDF ch. \emph{n}-grams and words & 0.626\\
SVM (CECL)&  BM25 ch. \emph{n}-grams 1-5 & 0.625\\
NB (CLUZH)& ch. \emph{n}-grams 2-6 & 0.616 \\
Cross Entropy (ahaqst) & ch. \emph{n}-grams up to 25 bytes & 0.614\\
Perplexity (Citius\_Ixa\_Imaxin) & words & 0.612\\
Perplexity (Citius\_Ixa\_Imaxin) & ch. 5-7 and word 1-3 \emph{n}-grams & 0.611\\
Naive Bayes (XAC\_Bayesline) & TF-IDF & 0.605\\
Perplexity (Citius\_Ixa\_Imaxin) & ch. 7-grams & 0.577\\
Cross Entropy (ahaqst) & word \emph{n}-grams 1-3 & 0.548\\
LSTM NN (deepCybErNet) & characters or words & 0.263\\
\hline
\end{tabular}
\caption{The weighted F1-scores using different methods on the 2017 GDI test set. The results from the experiments presented in this article are bolded.}
\label{GDI1list}
\end{table}

\subsection{Experiments on the GDI 2018 dataset}

\subsubsection{Baseline results and parameter estimation}

The GDI 2018 dataset included a separate development set (Table~\ref{setsizesGDI2018}). We used the development set to find the best parameters for the HeLI 2.0 method using the macro F1-score as the performance measure. The macro F1-score of 0.659 was obtained by the HeLI 2.0 method using just character \emph{n}-grams of the size 4 with \(p_{mod}=1.15\). The corresponding recall 66.17\% was slightly higher than the 66.10\% obtained with the HeLI method used in the GDI 2018 shared task. We then used the combined training and the development sets to train the language models. On the test set, the language identifier using these parameters obtained a macro F1-score of 0.650. The HeLI 2.0 method reached 0.011 higher macro F1-score than the HeLI method we used in the shared task. Even without the LM adaptation, the HeLI 2.0 method beats all the other reported methods.

\subsubsection{Experiments with language model adaptation}

Table~\ref{tab:koos2} shows the increment of the macro F1-score with different values of \(k\), the number of parts the examined mystery text collection is split into, on the development set using the same parameters with the HeLI 2.0 method as for the baseline. On the development set, \(k=57\) gave the best F1-score, with the absolute increase of 0.116 over the baseline. The corresponding recall was 77.74\%, which was somewhat lower than the 77.99\% obtained at the shared task.

\begin{table}[h]
\small
\center
\begin{tabular}{lc|lc}
\hline
\bf \emph{k} & \bf Macro F1-score & \bf \emph{k} & \bf Macro F1-score\\
\hline
1 & 0.659 & 52, 54, or 55 & 0.774\\
2 & 0.719 & 56 or 57 & 0.776\\
4 & 0.755 & 58 & 0.774\\
8 & 0.769 & 60 or 64 & 0.775\\
16 & 0.773 & 96 or 128 & 0.774\\
32, 40, 44, or 46 & 0.774 & 256 or 512 & 0.775\\
48 & 0.775 & 1024, 2048, or 4658 & 0.774\\
\hline
\end{tabular}
\caption{The macro F1-scores gained with different values of \(k\) when evaluated on the development set.}
\label{tab:koos2}
\end{table}

\subsubsection{Results of the LM adaptation on the test set}

Based on the evaluations using the development set, we decided to use \(k=57\) for the test run. All the training and the development data was used for the initial LM creation. The method using the LM adaptation algorithm reached the macro F1-score of 0.707. This macro F1-score is 0.057 higher than the one obtained by the non-adaptive version and 0.021 higher than results we obtained using language model adaptation in the GDI 2018 shared task.

\subsubsection{Iterative adaptation}

We tested repeating the LM adaptation algorithm for several epochs and the results of those trials on the GDI 2018 development set can be seen in Table~\ref{tab:epochs2}. There was a clear improvement of 0.041, at 477-999 epochs, on the original macro F1-score. It would again seem that, the language models become very stable with repeated adaptation, at least when there is no unknown language present in the data which is the case with the development set. Good scores were obtained already at 20 iterations, after which the results started to fluctuate up and down.

\begin{table}[h]
\small
\center
\begin{tabular}{lc|lc}
\hline
\bf Number of epochs & \bf Macro F1-score & \bf Number of epochs & \bf Macro F1-score\\
\hline
1 & 0.776& 17-19 & 0.813\\
2 & 0.787& 20 & 0.814\\
3 & 0.792& 21 & 0.813\\
4 & 0.797& 22-33 & 0.814\\
5 & 0.800& 34-54 & 0.815\\
6 & 0.801& 55-82 & 0.816\\
7 & 0.804& 83-88 & 0.815\\
8 & 0.806& 89-94 & 0.816\\
9 & 0.807& 95-111 & 0.815\\
10 & 0.808& 112-122 & 0.816\\
11 & 0.809& 123-129 & 0.815\\
12-13 & 0.810& 130-476 & 0.816\\
14 & 0.811& 477-999 & 0.817\\
15-16 & 0.812\\
\hline
\end{tabular}
\caption{Macro F1-scores with iterative LM adaptation on the GDI 2018 development set.}
\label{tab:epochs2}
\end{table}

Based on the results on the development set, we decided to try two different counts of iterations: 738, which is the number of epochs in the middle of the best scores, and 20, after which the results started to fluctuate. The tests resulted in a macro F1-score of 0.696 for 738 epochs and 0.704 for 20 epochs. As an additional experiment, we evaluated the iterative adaptation on a test set, from which the unknown dialects had been removed and obtained an F1-score of 0.729 with 738 epochs. From the results, it is clear that the presence of the unknown language is detrimental to repeated language model adaptation. In Table~\ref{GDI2list}, we report the macro F1-scores obtained by the teams participating in the GDI 2018 shared task, as well as our own. The methods used are listed in the first column, used features in the second column, and the best reached macro F1-score in the third column.

\begin{table}[h]
\small
\center
\begin{tabular}{lcc}
\hline
\bf Method (Team) & \bf Features used & \bf F1\\
\hline
\bf HeLI 2.0 with LM adapt. & ch. 4-grams & \bf 0.707\\
\bf HeLI 2.0 with iter. LM adapt. & ch. 4-grams & \bf 0.696\\
HeLI with LM adapt. (SUKI) & ch. 4-grams & 0.686\\
\bf HeLI 2.0 & ch. 4-grams & \bf 0.650\\
SVM ensemble (Twist Bytes) & ch. and word \emph{n}-grams 1-7 & 0.646\\
CNN with GRU (safina) & characters & 0.645\\
SVMs (T\"ubingen-Oslo) & ch. \emph{n}-grams 1-6, word \emph{n}-grams 1-3 & 0.640\\
HeLI (SUKI) & ch. 4-grams & 0.639\\
Voting ensemble (LaMa) & ch. \emph{n}-grams 1-8, word \emph{n}-grams 1-6 & 0.637\\
Na\"ive Bayes (XAC) & TF-IDF ch. \emph{n}-grams 1-6 & 0.634\\
Ridge regression (XAC) & TF-IDF ch. \emph{n}-grams 1-6 & 0.630\\
SGD (XAC) & TF-IDF ch. \emph{n}-grams 1-6 & 0.630\\
CNN (safina) & characters & 0.645\\
SVM ensemble (GDI\_classification) & ch. \emph{n}-grams 2-5 & 0.620\\
CNN with embedding (safina) & characters & 0.645\\
RNN with LSTM (T\"ubingen-Oslo) & & 0.616 \\
Euclidean distance (dkosmajac) & ch. \emph{n}-grams & 0.591\\
\hline
\end{tabular}
\caption{The macro F1-scores using different methods on the 2018 GDI test set. The results from the experiments presented in this article are bolded.}
\label{GDI2list}
\end{table}

\subsection{Experiments on the ILI 2018 dataset}

\subsubsection{Baseline results and parameter estimation}

We used the development set to find the best parameters for the HeLI 2.0 method using the macro F1-score as the measure. Using both original and lowercased character \emph{n}-grams from one to six with \(p_{mod}=1.09\), the method obtained the macro F1-score of 0.954. The corresponding recall was 95.26\%, which was exactly the same we obtained with the HeLI method used in the ILI 2018 shared task. We then used the combined training and the development sets to train the language models. On the test set, the language identifier using the above parameters obtained a macro F1-score of 0.880, which was clearly lower than the score we obtained using the HeLI method in the shared task.

\subsubsection{Experiments with language model adaptation}

Table~\ref{tab:koos3} shows the increment of the macro F1-score with different values of \(k\) on the development set using the same parameters with the HeLI 2.0 method as for the baseline. On the development set, \(k=64\) gave the best F1-score, 0.964, which is an absolute increase of 0.010 on the original F1-score. The corresponding recall was 96.29\%, which was a bit better than the 96.22\% obtained in the shared task.

\begin{table}[h]
\small
\center
\begin{tabular}{lc|lc}
\hline
\bf \emph{k} & \bf Macro F1-score & \bf \emph{k} & \bf Macro F1-score\\
\hline
1 & 0.954 & 32 or 48 & 0.964\\
2 & 0.958 & 58 & 0.963\\
4 & 0.960 & 60 or 62 - & 0.964\\
8 or 16 & 0.963\\
\hline
\end{tabular}
\caption{The macro F1-scores gained with different values of \(k\) when tested on the ILI 2018 development set.}
\label{tab:koos3}
\end{table}

\subsubsection{Results of the LM adaptation on the test data}

Based on the evaluations using the development data, we decided to use \(k=64\) as the number of splits for the actual test run. All the training and the development data was used for the initial LM creation. The identifier using the LM adaptation algorithm obtained a macro F1-score of 0.955. This macro F1-score is basically the same we obtained with language model adaptation in the ILI 2018 shared task, only some small fractions lower.

\subsubsection{Iterative adaptation}

We experimented repeating the LM adaptation algorithm for several epochs and the results of those trials on the development set can be seen in Table~\ref{tab:epochs3}. There was a a very small improvement of 0.001 on the original macro F1-score. The best absolute F-scores were reached at epochs 17 and 18. It would again seem that the language models become very stable with repeated adaptation.

\begin{table}[h]
\small
\center
\begin{tabular}{lc|lc}
\hline
\bf Number of epochs & \bf Macro F1-score & \bf Number of epochs & \bf Macro F1-score\\
\hline
1 & 0.964 &
2 - 999 & 0.965\\   
\hline
\end{tabular}
\caption{Macro F1-scores with iterative LM adaptation on the ILI 2018 development set.}
\label{tab:epochs3}
\end{table}

Based on the results on the development set, we decided to use LM adaptation with 18 iterations on the test set. The test resulted in a macro F1-score of 0.958, which is again almost the same as in the shared task, though this time some small fractions higher. We report the F1-scores obtained by the different teams participating in the ILI 2018 shared task in Table~\ref{ILI1list}, with the results form this article in bold. The methods used are listed in the first column, used features in the second column, and the macro F1-scores in the third column.

\begin{table}[h]
\small
\center
\begin{tabular}{lcc}
\hline
\bf Method (Team) & \bf Features used & \bf F1\\
\hline
\bf HeLI 2.0 with iter. LM adapt. & ch. \emph{n}-grams 1-6 & \bf 0.958 \\
HeLI with iter. LM adapt. (SUKI) & ch. \emph{n}-grams 1-6 & 0.958 \\
HeLI with LM adapt. (SUKI) & ch. \emph{n}-grams 1-6 & 0.955 \\
\bf HeLI 2.0 with LM adapt. & ch. \emph{n}-grams 1-6 & \bf 0.955 \\
SVM (T\"ubingen-Oslo) & ch. \emph{n}-grams 1-6, word \emph{n}-grams 1-3 & 0.902 \\
Ridge regression (XAC) & ch. \emph{n}-grams 2-6 & 0.898 \\
SVM ensemble (ILIdentification) &  ch. \emph{n}-grams 2-4 & 0.889 \\
HeLI (SUKI) & ch. \emph{n}-grams 1-6 & 0.887 \\
SGD (XAC) & ch. \emph{n}-grams 2-6 & 0.883 \\
\bf HeLI 2.0 & ch. \emph{n}-grams 1-6 & \bf 0.880\\
CNN (safina) & characters & 0.863 \\
NB (XAC) & ch. \emph{n}-grams 2-6 & 0.854 \\
Euclidean distance (dkosmajac) & & 0.847 \\
CNN with embedding (safina) & characters & 0.863 \\
LSTM RNN (we\_are\_indian) & words & 0.836 \\
CNN with GRU (safina) & characters & 0.826 \\
NB (LaMa) & & 0.819 \\
RNN with GRU (T\"ubingen-Oslo) & & 0.753 \\
Mutual information (we\_are\_indian) & & 0.744 \\
\hline
\end{tabular}
\caption{The macro F1-scores using different methods on the 2018 ILI test set. The results presented for the first time are in bold.}
\label{ILI1list}
\end{table}

\section{Discussion}

The 26\% difference in F1-scores between the development portion (0.890) and the test set (0.659) of the GDI 2017 data obtained by the HeLI 2.0 method is considerable. It seems to indicate that the test set contains more out-of-domain material when compared with the partition of the training set we used for development. In order to validate this hypothesis, we divided the test set into two parts. The 2nd part remained to be used for testing in four scenarios with the HeLI 2.0 method. In the scenarios we used different combinations of data for training: the original training set, the training set augmented with the first part of test data, the training set of which a part was replaced by the first part of the test set, and only using the first part of the test set. The results of these experiments support our hypothesis, as can be seen in Table~\ref{tab:devtest}. The domain difference between the two sets explains why iterative adaptation performs better with the test set than with the development set. After each iteration, the relative amount of the original training data gets smaller, as the information from the test data is repeatedly added to the language models.

\begin{table}[h]
\small
\center
\begin{tabular}{lc}
\hline
\bf Data used for the language models & \bf Macro F1\\
\hline
training set  & 0.656\\
training set + 1st part of test set & 0.801\\
part of training set replaced with 1st part of test set & 0.803\\
1st part of test set & 0.858\\
\hline
\end{tabular}
\caption{The macro F1-scores for the second part of test set using different training data combinations.}
\label{tab:devtest}
\end{table}

In the GDI 2018 dataset, there is only a 1.4\% difference between the macro F1-scores obtained from the development and the test sets. This indicates that the GDI 2018 development set is in the same way out-of-domain when compared with the training set as the actual test set is.

There is a small difference (7.8\%) between the F1-scores attained using the development set and the test set of the ILI 2018 data as well. However, such small differences can be partly due to the fact that the parameters of the identification method have been optimized using the development set. 

\section{Conclusions}

The results indicate that unsupervised LM adaptation should be considered in all language identification tasks, especially in those where the amount of out-of-domain data is significant. If the presence of unseen languages is to be expected, the use of language model adaptation could still be beneficial, but special care must be taken as repeated adaptation in particular could decrease the identification accuracy.

Though the iterative LM adaptation is computationally costly when compared with the baseline HeLI 2.0 method, it must be noted that the final identifications with 485 epochs on the GDI 2017 test set took only around 20 minutes using one computing core of a modern laptop.

\section*{Acknowledgments}

This research was partly conducted with funding from the Kone Foundation Language Programme \cite{kone1}.

\bibliography{strings,LanguageIdentification}

\begin{thebibliography}{}

\bibitem[\protect\BCAY{Ali}{Ali}{2018a}]{aliGDI2018}
Ali, M. \BBOP2018a\BBCP.
\newblock \BBOQ Character level convolutional neural network for german dialect
  identification\BBCQ\
\newblock In {\Bem Proceedings of the Fifth Workshop on NLP for Similar
  Languages, Varieties and Dialects (VarDial 2018)}, \BPGS\ 172--177.

\bibitem[\protect\BCAY{Ali}{Ali}{2018b}]{aliILI2018}
Ali, M. \BBOP2018b\BBCP.
\newblock \BBOQ Character level convolutional neural network for indo-aryan
  language identification\BBCQ\
\newblock In {\Bem Proceedings of the Fifth Workshop on NLP for Similar
  Languages, Varieties and Dialects (VarDial 2018)}, \BPGS\ 283--287.

\bibitem[\protect\BCAY{Bacchiani\ \BBA\ Roark}{Bacchiani\ \BBA\
  Roark}{2003}]{bacchiani1}
Bacchiani, M.\BBACOMMA\  \BBA\ Roark, B. \BBOP2003\BBCP.
\newblock \BBOQ Unsupervised language model adaptation\BBCQ\
\newblock In {\Bem 2003 IEEE International Conference on Acoustics, Speech, and
  Signal Processing, 2003. Proceedings.(ICASSP'03).}, \lowercase{\BVOL}~1,
  \BPGS\ I--I. IEEE.

\bibitem[\protect\BCAY{Barbaresi}{Barbaresi}{2017}]{barbaresi:2017:VarDial}
Barbaresi, A. \BBOP2017\BBCP.
\newblock \BBOQ Discriminating between similar languages using weighted subword
  features\BBCQ\
\newblock In {\Bem Proceedings of the Fourth Workshop on NLP for Similar
  Languages, Varieties and Dialects (VarDial)}, \BPGS\ 184--189, Valencia,
  Spain.

\bibitem[\protect\BCAY{Barbaresi}{Barbaresi}{2018}]{barbaresiGDI2018}
Barbaresi, A. \BBOP2018\BBCP.
\newblock \BBOQ Computationally efficient discrimination between language
  varieties with large feature vectors and regularized classifiers\BBCQ\
\newblock In {\Bem Proceedings of the Fifth Workshop on NLP for Similar
  Languages, Varieties and Dialects (VarDial 2018)}, \BPGS\ 164--171.

\bibitem[\protect\BCAY{Benites, Grubenmann, von D{\"a}niken, von Gr{\"u}nigen,
  Deriu,\ \BBA\ Cieliebak}{Benites et~al.}{2018}]{benitesGDI2018}
Benites, F., Grubenmann, R., von D{\"a}niken, P., von Gr{\"u}nigen, D., Deriu,
  J., \BBA\ Cieliebak, M. \BBOP2018\BBCP.
\newblock \BBOQ Twist bytes-german dialect identification with data mining
  optimization\BBCQ\
\newblock In {\Bem Proceedings of the Fifth Workshop on NLP for Similar
  Languages, Varieties and Dialects (VarDial 2018)}, \BPGS\ 218--227.

\bibitem[\protect\BCAY{Bergsma, McNamee, Bagdouri, Fink,\ \BBA\ Wilson}{Bergsma
  et~al.}{2012}]{bergsma1}
Bergsma, S., McNamee, P., Bagdouri, M., Fink, C., \BBA\ Wilson, T.
  \BBOP2012\BBCP.
\newblock \BBOQ {Language Identification for Creating Language-specific Twitter
  Collections}\BBCQ\
\newblock In {\Bem Proceedings of the Second Workshop on Language in Social
  Media (LSM2012)}, \BPGS\ 65--74, Montréal, Canada.

\bibitem[\protect\BCAY{Bestgen}{Bestgen}{2017}]{bestgen:2017:VarDial}
Bestgen, Y. \BBOP2017\BBCP.
\newblock \BBOQ Improving the character ngram model for the dsl task with bm25
  weighting and less frequently used feature sets\BBCQ\
\newblock In {\Bem Proceedings of the Fourth Workshop on NLP for Similar
  Languages, Varieties and Dialects (VarDial)}, \BPGS\ 115--123, Valencia,
  Spain.

\bibitem[\protect\BCAY{Blodgett, Wei,\ \BBA\ O'Connor}{Blodgett
  et~al.}{2017}]{blodgett1}
Blodgett, S.~L., Wei, J. T.-Z., \BBA\ O'Connor, B. \BBOP2017\BBCP.
\newblock \BBOQ {A Dataset and Classifier for Recognizing Social Media
  English}\BBCQ\
\newblock In {\Bem Proceedings of the 3rd Workshop on Noisy User-generated
  Text}, \BPGS\ 56--61, Copenhagen, Denmark.

\bibitem[\protect\BCAY{Brown}{Brown}{2012}]{brown1}
Brown, R.~D. \BBOP2012\BBCP.
\newblock \BBOQ {Finding and Identifying Text in 900+ Languages}\BBCQ\
\newblock {\Bem Digital Investigation}, {\Bem 9}, S34--S43.

\bibitem[\protect\BCAY{Brown}{Brown}{2013}]{brown2}
Brown, R.~D. \BBOP2013\BBCP.
\newblock \BBOQ {Selecting and Weighting N-grams to Identify 1100
  Languages}\BBCQ\
\newblock In {\Bem Proceedings of the 16th International Conference on Text,
  Speech and Dialogue (TSD 2013)}, \BPGS\ 475--483, Plze\v{n}, Czech Republic.

\bibitem[\protect\BCAY{Brown}{Brown}{2014}]{brown3}
Brown, R.~D. \BBOP2014\BBCP.
\newblock \BBOQ {Non-linear Mapping for Improved Identification of 1300+
  Languages}\BBCQ\
\newblock In {\Bem Proceedings of the 2014 Conference on Empirical Methods in
  Natural Language Processing (EMNLP 2014)}, \BPGS\ 627--632, Doha, Qatar.

\bibitem[\protect\BCAY{Chen\ \BBA\ Maison}{Chen\ \BBA\ Maison}{2003}]{chen1}
Chen, S.~F.\BBACOMMA\  \BBA\ Maison, B. \BBOP2003\BBCP.
\newblock \BBOQ {Using Place Name Data to Train Language Identification
  Models}\BBCQ\
\newblock In {\Bem 8th European Conference on Speech Communication and
  Technology EUROSPEECH 2003 - INTERSPEECH 2003}, \BPGS\ 1349--1352, Geneva,
  Switzerland.

\bibitem[\protect\BCAY{Ciobanu, Malmasi,\ \BBA\ Dinu}{Ciobanu
  et~al.}{2018a}]{ciobanuGDI2018}
Ciobanu, A.~M., Malmasi, S., \BBA\ Dinu, L.~P. \BBOP2018a\BBCP.
\newblock \BBOQ German dialect identification using classifier ensembles\BBCQ\
\newblock {\Bem arXiv preprint arXiv:1807.08230}.

\bibitem[\protect\BCAY{Ciobanu, Zampieri, Malmasi, Pal,\ \BBA\ Dinu}{Ciobanu
  et~al.}{2018b}]{ciobanuILI2018}
Ciobanu, A.~M., Zampieri, M., Malmasi, S., Pal, S., \BBA\ Dinu, L.~P.
  \BBOP2018b\BBCP.
\newblock \BBOQ Discriminating between indo-aryan languages using svm
  ensembles\BBCQ\
\newblock {\Bem arXiv preprint arXiv:1807.03108}.

\bibitem[\protect\BCAY{Clematide\ \BBA\ Makarov}{Clematide\ \BBA\
  Makarov}{2017}]{clematide-makarov:2017:VarDial}
Clematide, S.\BBACOMMA\  \BBA\ Makarov, P. \BBOP2017\BBCP.
\newblock \BBOQ Cluzh at vardial gdi 2017: Testing a variety of machine
  learning tools for the classification of swiss german dialects\BBCQ\
\newblock In {\Bem Proceedings of the Fourth Workshop on NLP for Similar
  Languages, Varieties and Dialects (VarDial)}, \BPGS\ 170--177, Valencia,
  Spain.

\bibitem[\protect\BCAY{Coltekin\ \BBA\ Rama}{Coltekin\ \BBA\
  Rama}{2017}]{coltekin2}
Coltekin, C.\BBACOMMA\  \BBA\ Rama, T. \BBOP2017\BBCP.
\newblock \BBOQ {Tübingen System in VarDial 2017 Shared Task: Experiments with
  Language Identification and Cross-lingual Parsing}\BBCQ\
\newblock In {\Bem Proceedings of the Fourth Workshop on NLP for Similar
  Languages, Varieties and Dialects}, \BPGS\ 146--155, Valencia, Spain.

\bibitem[\protect\BCAY{Coltekin, Rama,\ \BBA\ Blaschke}{Coltekin
  et~al.}{2018}]{coltekinGDI2018}
Coltekin, C., Rama, T., \BBA\ Blaschke, V. \BBOP2018\BBCP.
\newblock \BBOQ T{\"u}bingen-oslo team at the vardial 2018 evaluation campaign:
  An analysis of n-gram features in language variety identification\BBCQ\
\newblock In {\Bem Proceedings of the Fifth Workshop on NLP for Similar
  Languages, Varieties and Dialects (VarDial 2018)}, \BPGS\ 55--65.

\bibitem[\protect\BCAY{Gamallo, Pichel,\ \BBA\ Alegria}{Gamallo
  et~al.}{2017}]{gamallo-pichel-alegria:2017:VarDial}
Gamallo, P., Pichel, J.~R., \BBA\ Alegria, I.~n. \BBOP2017\BBCP.
\newblock \BBOQ A perplexity-based method for similar languages
  discrimination\BBCQ\
\newblock In {\Bem Proceedings of the Fourth Workshop on NLP for Similar
  Languages, Varieties and Dialects (VarDial)}, \BPGS\ 109--114, Valencia,
  Spain.

\bibitem[\protect\BCAY{Gupta, Dhakad, Gupta,\ \BBA\ Singh}{Gupta
  et~al.}{2018}]{guptaILI2018}
Gupta, D., Dhakad, G., Gupta, J., \BBA\ Singh, A.~K. \BBOP2018\BBCP.
\newblock \BBOQ Iit (bhu) system for indo-aryan language identification (ili)
  at vardial 2018\BBCQ\
\newblock In {\Bem Proceedings of the Fifth Workshop on NLP for Similar
  Languages, Varieties and Dialects (VarDial 2018)}, \BPGS\ 185--190.

\bibitem[\protect\BCAY{Hanani, Qaroush,\ \BBA\ Taylor}{Hanani
  et~al.}{2017}]{hanani-qaroush-taylor:2017:VarDial}
Hanani, A., Qaroush, A., \BBA\ Taylor, S. \BBOP2017\BBCP.
\newblock \BBOQ Identifying dialects with textual and acoustic cues\BBCQ\
\newblock In {\Bem Proceedings of the Fourth Workshop on NLP for Similar
  Languages, Varieties and Dialects (VarDial)}, \BPGS\ 93--101, Valencia,
  Spain.

\bibitem[\protect\BCAY{Hasimu\ \BBA\ Silamu}{Hasimu\ \BBA\
  Silamu}{2018}]{hasimu3}
Hasimu, M.\BBACOMMA\  \BBA\ Silamu, W. \BBOP2018\BBCP.
\newblock \BBOQ {On Hierarchical Text Language-Identification Algorithms}\BBCQ\
\newblock {\Bem Algorithms}, {\Bem 11\/}(39).

\bibitem[\protect\BCAY{Hollenstein\ \BBA\ Aepli}{Hollenstein\ \BBA\
  Aepli}{2015}]{hollenstein1}
Hollenstein, N.\BBACOMMA\  \BBA\ Aepli, N. \BBOP2015\BBCP.
\newblock \BBOQ {A Resource for Natural Language Processing of Swiss German
  Dialects}\BBCQ\
\newblock In {\Bem Proceedings of GSCL}, \BPGS\ 108--109.

\bibitem[\protect\BCAY{Hosmer, Lemeshow,\ \BBA\ Sturdivant}{Hosmer
  et~al.}{2013}]{hosmer1}
Hosmer, D.~W., Lemeshow, S., \BBA\ Sturdivant, R.~X. \BBOP2013\BBCP.
\newblock {\Bem {Applied logistic regression}\/} (3rd ed \BEd).
\newblock Wiley Series in Probability and Statistics. Wiley, Hoboken, N.J.,
  USA.

\bibitem[\protect\BCAY{Indhuja, Indu, Sreejith,\ \BBA\ Reghu~Raj}{Indhuja
  et~al.}{2014}]{indhuja2}
Indhuja, K., Indu, M., Sreejith, C., \BBA\ Reghu~Raj, P.~C. \BBOP2014\BBCP.
\newblock \BBOQ {Text Based Language Identification System for Indian Languages
  Following Devanagiri Script}\BBCQ\
\newblock {\Bem International Journal of Engineering Reseach and Technology},
  {\Bem 3(4)}, 327--331.

\bibitem[\protect\BCAY{Ionescu}{Ionescu}{2013}]{ionescu2}
Ionescu, R.~T. \BBOP2013\BBCP.
\newblock \BBOQ Local rank distance\BBCQ\
\newblock In Björner, N., Negru, V., Ida, T., Jebelean, T., Petcu, D., Watt,
  S., \BBA\ Zaharie, D.\BEDS, {\Bem {Proceedings of the 15th International
  Symposium on Symbolic and Numeric Algorithms for Scientific Computing (SYNASC
  2013)}}, \BPGS\ 219--226, Timisoara, Romania.

\bibitem[\protect\BCAY{Ionescu\ \BBA\ Butnaru}{Ionescu\ \BBA\
  Butnaru}{2017}]{ionescu-butnaru:2017:VarDial}
Ionescu, R.~T.\BBACOMMA\  \BBA\ Butnaru, A. \BBOP2017\BBCP.
\newblock \BBOQ Learning to identify arabic and german dialects using multiple
  kernels\BBCQ\
\newblock In {\Bem Proceedings of the Fourth Workshop on NLP for Similar
  Languages, Varieties and Dialects (VarDial)}, \BPGS\ 200--209, Valencia,
  Spain.

\bibitem[\protect\BCAY{Jauhiainen, Jauhiainen,\ \BBA\ Lind{\'e}n}{Jauhiainen
  et~al.}{2015}]{jauhiainen4}
Jauhiainen, H., Jauhiainen, T., \BBA\ Lind{\'e}n, K. \BBOP2015\BBCP.
\newblock \BBOQ {The Finno-Ugric Languages and The Internet Project}\BBCQ\
\newblock {\Bem Septentrio Conference Series}, {\Bem 0\/}(2), 87--98.

\bibitem[\protect\BCAY{Jauhiainen}{Jauhiainen}{2010}]{jauhiainen1}
Jauhiainen, T. \BBOP2010\BBCP.
\newblock \BBOQ Tekstin kielen automaattinen tunnistaminen\BBCQ\
\newblock Master's thesis, University of Helsinki, Helsinki.

\bibitem[\protect\BCAY{Jauhiainen, Jauhiainen,\ \BBA\ Lind\'{e}n}{Jauhiainen
  et~al.}{2015}]{jauhiainen-jauhiainen-linden:2015:LT4VarDial}
Jauhiainen, T., Jauhiainen, H., \BBA\ Lind\'{e}n, K. \BBOP2015\BBCP.
\newblock \BBOQ Discriminating similar languages with token-based backoff\BBCQ\
\newblock In {\Bem Proceedings of the Joint Workshop on Language Technology for
  Closely Related Languages, Varieties and Dialects (LT4VarDial)}, \BPGS\
  44--51, Hissar, Bulgaria.

\bibitem[\protect\BCAY{Jauhiainen, Jauhiainen,\ \BBA\ Lind\'{e}n}{Jauhiainen
  et~al.}{2018a}]{jauhiainen-linden-jauhiainen:2018:DFS}
Jauhiainen, T., Jauhiainen, H., \BBA\ Lind\'{e}n, K. \BBOP2018a\BBCP.
\newblock \BBOQ {HeLI-based Experiments in Discriminating Between Dutch and
  Flemish Subtitles}\BBCQ\
\newblock In {\Bem Proceedings of the Fifth Workshop on NLP for Similar
  Languages, Varieties and Dialects (VarDial)}, \BPGS\ 137--144, Santa Fe, NM.

\bibitem[\protect\BCAY{Jauhiainen, Jauhiainen,\ \BBA\ Lind\'{e}n}{Jauhiainen
  et~al.}{2018b}]{jauhiainen-linden-jauhiainen:2018:GDI}
Jauhiainen, T., Jauhiainen, H., \BBA\ Lind\'{e}n, K. \BBOP2018b\BBCP.
\newblock \BBOQ {HeLI-based Experiments in Swiss German Dialect
  Identification}\BBCQ\
\newblock In {\Bem Proceedings of the Fifth Workshop on NLP for Similar
  Languages, Varieties and Dialects (VarDial)}, \BPGS\ 254--262, Santa Fe, NM.

\bibitem[\protect\BCAY{Jauhiainen, Jauhiainen,\ \BBA\ Lind\'{e}n}{Jauhiainen
  et~al.}{2018c}]{jauhiainen-linden-jauhiainen:2018:ILI}
Jauhiainen, T., Jauhiainen, H., \BBA\ Lind\'{e}n, K. \BBOP2018c\BBCP.
\newblock \BBOQ {Iterative Language Model Adaptation for Indo-Aryan Language
  Identification}\BBCQ\
\newblock In {\Bem Proceedings of the Fifth Workshop on NLP for Similar
  Languages, Varieties and Dialects (VarDial)}, \BPGS\ 66--75, Santa Fe, NM.

\bibitem[\protect\BCAY{Jauhiainen, Lind{\'e}n,\ \BBA\ Jauhiainen}{Jauhiainen
  et~al.}{2015}]{jauhiainen3}
Jauhiainen, T., Lind{\'e}n, K., \BBA\ Jauhiainen, H. \BBOP2015\BBCP.
\newblock \BBOQ {Language Set Identification in Noisy Synthetic Multilingual
  Documents}\BBCQ\
\newblock In {\Bem Proceedings of the Computational Linguistics and Intelligent
  Text Processing 16th International Conference, CICLing 2015}, \BPGS\
  633--643, Cairo, Egypt.

\bibitem[\protect\BCAY{Jauhiainen, Lind\'{e}n,\ \BBA\ Jauhiainen}{Jauhiainen
  et~al.}{2016}]{jauhiainen-linden-jauhiainen:2016:VarDial3}
Jauhiainen, T., Lind\'{e}n, K., \BBA\ Jauhiainen, H. \BBOP2016\BBCP.
\newblock \BBOQ {HeLI, a Word-Based Backoff Method for Language
  Identification}\BBCQ\
\newblock In {\Bem Proceedings of the Third Workshop on NLP for Similar
  Languages, Varieties and Dialects (VarDial3)}, \BPGS\ 153--162, Osaka, Japan.

\bibitem[\protect\BCAY{Jauhiainen, Lind\'{e}n,\ \BBA\ Jauhiainen}{Jauhiainen
  et~al.}{2017a}]{jauhiainen-linden-jauhiainen:2017:VarDial}
Jauhiainen, T., Lind\'{e}n, K., \BBA\ Jauhiainen, H. \BBOP2017a\BBCP.
\newblock \BBOQ Evaluating heli with non-linear mappings\BBCQ\
\newblock In {\Bem Proceedings of the Fourth Workshop on NLP for Similar
  Languages, Varieties and Dialects (VarDial)}, \BPGS\ 102--108, Valencia,
  Spain.

\bibitem[\protect\BCAY{Jauhiainen, Lind{\'e}n,\ \BBA\ Jauhiainen}{Jauhiainen
  et~al.}{2017b}]{jauhiainen6}
Jauhiainen, T., Lind{\'e}n, K., \BBA\ Jauhiainen, H. \BBOP2017b\BBCP.
\newblock \BBOQ {Evaluation of Language Identification Methods Using 285
  Languages}\BBCQ\
\newblock In {\Bem Proceedings of the 21st Nordic Conference on Computational
  Linguistics (NoDaLiDa 2017)}, \BPGS\ 183--191, Gothenburg, Sweden.
  Link{\"o}ping University Electronic Press.

\bibitem[\protect\BCAY{Jauhiainen, Lui, Zampieri, Baldwin,\ \BBA\
  Lind{\'e}n}{Jauhiainen et~al.}{2018}]{jauhiainen2018automatic}
Jauhiainen, T., Lui, M., Zampieri, M., Baldwin, T., \BBA\ Lind{\'e}n, K.
  \BBOP2018\BBCP.
\newblock \BBOQ Automatic language identification in texts: A survey\BBCQ\
\newblock {\Bem arXiv preprint arXiv:1804.08186}.

\bibitem[\protect\BCAY{Jelinek, Merialdo, Roukos,\ \BBA\ Strauss}{Jelinek
  et~al.}{1991}]{jelinek1}
Jelinek, F., Merialdo, B., Roukos, S., \BBA\ Strauss, M. \BBOP1991\BBCP.
\newblock \BBOQ A dynamic language model for speech recognition\BBCQ\
\newblock In {\Bem Speech and Natural Language: Proceedings of a Workshop Held
  at Pacific Grove, California, February 19-22, 1991}.

\bibitem[\protect\BCAY{{Kone Foundation}}{{Kone Foundation}}{2012}]{kone1}
{Kone Foundation} \BBOP2012\BBCP.
\newblock \BBOQ The language programme 2012-2016\BBCQ\
\newblock http://www.koneensaatio.fi/en.

\bibitem[\protect\BCAY{Kruengkrai, Sornlertlamvanich,\ \BBA\
  Isahara}{Kruengkrai et~al.}{2006}]{kruengkrai1}
Kruengkrai, C., Sornlertlamvanich, V., \BBA\ Isahara, H. \BBOP2006\BBCP.
\newblock \BBOQ {Language, Script, and Encoding Identification with String
  Kernel Classifiers}\BBCQ\
\newblock In {\Bem Proceedings of the 1st International Conference on
  Knowledge, Information and Creativity Support Systems (KICSS 2006)},
  Ayutthaya, Thailand.

\bibitem[\protect\BCAY{Kumar, Lahiri, Alok, Ojha, Jain, Basit,\ \BBA\
  Dawar}{Kumar et~al.}{2018}]{KUMAR18.26}
Kumar, R., Lahiri, B., Alok, D., Ojha, A.~K., Jain, M., Basit, A., \BBA\ Dawar,
  Y. \BBOP2018\BBCP.
\newblock \BBOQ {Automatic Identification of Closely-related Indian Languages:
  Resources and Experiments}\BBCQ\
\newblock In {\Bem Proceedings of the Eleventh International Conference on
  Language Resources and Evaluation (LREC)}.

\bibitem[\protect\BCAY{Lewis, Simons,\ \BBA\ Fennig}{Lewis
  et~al.}{2013}]{lewis1}
Lewis, M.~P., Simons, G.~F., \BBA\ Fennig, C.~D.\BEDS. \BBOP2013\BBCP.
\newblock {\Bem Ethnologue: Languages of the world, seventeenth edition}.
\newblock SIL International, Dallas, Texas.

\bibitem[\protect\BCAY{Li, Cohn,\ \BBA\ Baldwin}{Li et~al.}{2018}]{li3}
Li, Y., Cohn, T., \BBA\ Baldwin, T. \BBOP2018\BBCP.
\newblock \BBOQ What's in a domain? learning domain-robust text representations
  using adversarial training\BBCQ\
\newblock In {\Bem Proceedings of the 2018 Conference of the North American
  Chapter of the Association for Computational Linguistics --- Human Language
  Technologies (NAACL HLT 2018)}, \BPGS\ 474--479, New Orleans, USA.

\bibitem[\protect\BCAY{Malmasi\ \BBA\ Zampieri}{Malmasi\ \BBA\
  Zampieri}{2017}]{malmasi-zampieri:2017:VarDial1}
Malmasi, S.\BBACOMMA\  \BBA\ Zampieri, M. \BBOP2017\BBCP.
\newblock \BBOQ German dialect identification in interview transcriptions\BBCQ\
\newblock In {\Bem Proceedings of the Fourth Workshop on NLP for Similar
  Languages, Varieties and Dialects (VarDial)}, \BPGS\ 164--169, Valencia,
  Spain.

\bibitem[\protect\BCAY{Malmasi, Zampieri, Ljube\v{s}i\'{c}, Nakov, Ali,\ \BBA\
  Tiedemann}{Malmasi et~al.}{2016}]{malmasi9}
Malmasi, S., Zampieri, M., Ljube\v{s}i\'{c}, N., Nakov, P., Ali, A., \BBA\
  Tiedemann, J. \BBOP2016\BBCP.
\newblock \BBOQ {Discriminating Between Similar Languages and Arabic Dialect
  Identification: A Report on the Third DSL Shared Task}\BBCQ\
\newblock In {\Bem Proceedings of the Third Workshop on NLP for Similar
  Languages, Varieties and Dialects}, \BPGS\ 1--14, Osaka, Japan.

\bibitem[\protect\BCAY{Medvedeva, Kroon,\ \BBA\ Plank}{Medvedeva
  et~al.}{2017}]{medvedeva1}
Medvedeva, M., Kroon, M., \BBA\ Plank, B. \BBOP2017\BBCP.
\newblock \BBOQ {When Sparse Traditional Models Outperform Dense Neural
  Networks: the Curious Case of Discriminating between Similar Languages}\BBCQ\
\newblock In {\Bem Proceedings of the Fourth Workshop on NLP for Similar
  Languages, Varieties and Dialects}, \BPGS\ 156--163, Valencia, Spain.

\bibitem[\protect\BCAY{Murthy\ \BBA\ Kumar}{Murthy\ \BBA\
  Kumar}{2006}]{murthy1}
Murthy, K.~N.\BBACOMMA\  \BBA\ Kumar, G.~B. \BBOP2006\BBCP.
\newblock \BBOQ {Language Identification from Small Text Samples}\BBCQ\
\newblock {\Bem Journal of Quantitative Linguistics}, {\Bem 13\/}(1), 57--80.

\bibitem[\protect\BCAY{Mustonen}{Mustonen}{1965}]{mustonen1}
Mustonen, S. \BBOP1965\BBCP.
\newblock \BBOQ {Multiple Discriminant Analysis in Linguistic Problems}\BBCQ\
\newblock {\Bem Statistical Methods in Linguistics}, {\Bem 4}, 37--44.

\bibitem[\protect\BCAY{Pla\ \BBA\ Hurtado}{Pla\ \BBA\ Hurtado}{2017}]{pla2}
Pla, F.\BBACOMMA\  \BBA\ Hurtado, L.-F. \BBOP2017\BBCP.
\newblock \BBOQ {Language Identification of Multilingual Posts from Twitter: A
  Case Study}\BBCQ\
\newblock {\Bem Knowledge and Information Systems}, {\Bem 51\/}(3), 965--989.

\bibitem[\protect\BCAY{Rangel, Rosso, Montes-y G{\'o}mez, Potthast,\ \BBA\
  Stein}{Rangel et~al.}{2018}]{kestemont1}
Rangel, F., Rosso, P., Montes-y G{\'o}mez, M., Potthast, M., \BBA\ Stein, B.
  \BBOP2018\BBCP.
\newblock \BBOQ Overview of the 6th author profiling task at pan 2018:
  multimodal gender identification in twitter\BBCQ\
\newblock {\Bem Working Notes Papers of the CLEF}.

\bibitem[\protect\BCAY{Rani, Ojha,\ \BBA\ Jha}{Rani et~al.}{2018}]{RANI18.16}
Rani, P., Ojha, A.~K., \BBA\ Jha, G.~N. \BBOP2018\BBCP.
\newblock \BBOQ Automatic language identification system for hindi and
  magahi\BBCQ\
\newblock {\Bem arXiv preprint arXiv:1804.05095}.

\bibitem[\protect\BCAY{Samard\v{z}i\'{c}, Scherrer,\ \BBA\
  Glaser}{Samard\v{z}i\'{c} et~al.}{2016}]{samardzic2016archimob}
Samard\v{z}i\'{c}, T., Scherrer, Y., \BBA\ Glaser, E. \BBOP2016\BBCP.
\newblock \BBOQ {ArchiMob} -- a corpus of spoken {S}wiss {G}erman\BBCQ\
\newblock In {\Bem Proceedings of LREC}.

\bibitem[\protect\BCAY{Scherrer\ \BBA\ Rambow}{Scherrer\ \BBA\
  Rambow}{2010}]{scherrer1}
Scherrer, Y.\BBACOMMA\  \BBA\ Rambow, O. \BBOP2010\BBCP.
\newblock \BBOQ {Word-based Dialect Identification with Georeferenced
  Rules}\BBCQ\
\newblock In {\Bem Proceedings of the 2010 Conference on Empirical Methods in
  Natural Language Processing (EMNLP 2010)}, \BPGS\ 1151--1161, Massachusetts,
  USA. Association for Computational Linguistics.

\bibitem[\protect\BCAY{Sibun\ \BBA\ Reynar}{Sibun\ \BBA\ Reynar}{1996}]{sibun1}
Sibun, P.\BBACOMMA\  \BBA\ Reynar, J.~C. \BBOP1996\BBCP.
\newblock \BBOQ {Language Identification: Examining the Issues}\BBCQ\
\newblock In {\Bem Proceedings of the 5th Annual Symposium on Document Analysis
  and Information Retrieval (SDAIR-96)}, \BPGS\ 125--135, Las Vegas, USA.

\bibitem[\protect\BCAY{Sreejith, Indu,\ \BBA\ Reghu~Raj}{Sreejith
  et~al.}{2013}]{sreejith1}
Sreejith, C., Indu, M., \BBA\ Reghu~Raj, P.~C. \BBOP2013\BBCP.
\newblock \BBOQ {N-gram based Algorithm for Distinguishing Between Hindi and
  Sanskrit Texts}\BBCQ\
\newblock In {\Bem Proceedings of the Fourth IEEE International Conference on
  Computing, Communication and Networking Technologies}, Tiruchengode, India.

\bibitem[\protect\BCAY{Tiedemann\ \BBA\ Ljube\v{s}i\'c}{Tiedemann\ \BBA\
  Ljube\v{s}i\'c}{2012}]{tiedemann1}
Tiedemann, J.\BBACOMMA\  \BBA\ Ljube\v{s}i\'c, N. \BBOP2012\BBCP.
\newblock \BBOQ {Efficient Discrimination Between Closely Related
  Languages}\BBCQ\
\newblock In {\Bem Proceedings of the 24th International Conference on
  Computational Linguistics (COLING 2012)}, \BPGS\ 2619--2634, Mumbai, India.

\bibitem[\protect\BCAY{Vaid\ \BBA\ Gupta}{Vaid\ \BBA\ Gupta}{2002}]{vaid1}
Vaid, J.\BBACOMMA\  \BBA\ Gupta, A. \BBOP2002\BBCP.
\newblock \BBOQ Exploring word recognition in a semi-alphabetic script: The
  case of devanagari\BBCQ\
\newblock {\Bem Brain and Language}, {\Bem 81\/}(1-3), 679--690.

\bibitem[\protect\BCAY{Vatanen, Väyrynen,\ \BBA\ Virpioja}{Vatanen
  et~al.}{2010}]{vatanen1}
Vatanen, T., Väyrynen, J.~J., \BBA\ Virpioja, S. \BBOP2010\BBCP.
\newblock \BBOQ {Language Identification of Short Text Segments with N-gram
  Models}\BBCQ\
\newblock In {\Bem Proceedings of the 7th International Conference on Language
  Resources and Evaluation (LREC 2010)}, \BPGS\ 3423--3430, Valletta, Malta.

\bibitem[\protect\BCAY{Zampieri, Malmasi, Ljubešic, Nakov, Ali, Tiedemann,
  Scherrer,\ \BBA\ Aepli}{Zampieri et~al.}{2017}]{zampieri9}
Zampieri, M., Malmasi, S., Ljubešic, N., Nakov, P., Ali, A., Tiedemann, J.,
  Scherrer, Y., \BBA\ Aepli, N. \BBOP2017\BBCP.
\newblock \BBOQ {Findings of the VarDial Evaluation Campaign 2017}\BBCQ\
\newblock In {\Bem Proceedings of the Fourth Workshop on NLP for Similar
  Languages, Varieties and Dialects}, \BPGS\ 1--15, Valencia, Spain.

\bibitem[\protect\BCAY{Zampieri, Malmasi, Nakov, Ali, Shuon, Glass, Scherrer,
  Samard\v{z}i{\'c}, Ljube\v{s}i\'{c}, Tiedemann, {van der Lee}, Grondelaers,
  Oostdijk, {van den Bosch}, Kumar, Lahiri,\ \BBA\ Jain}{Zampieri
  et~al.}{2018}]{vardial2018report}
Zampieri, M., Malmasi, S., Nakov, P., Ali, A., Shuon, S., Glass, J., Scherrer,
  Y., Samard\v{z}i{\'c}, T., Ljube\v{s}i\'{c}, N., Tiedemann, J., {van der
  Lee}, C., Grondelaers, S., Oostdijk, N., {van den Bosch}, A., Kumar, R.,
  Lahiri, B., \BBA\ Jain, M. \BBOP2018\BBCP.
\newblock \BBOQ {Language Identification and Morphosyntactic Tagging: The
  Second VarDial Evaluation Campaign}\BBCQ\
\newblock In {\Bem Proceedings of the Fifth Workshop on NLP for Similar
  Languages, Varieties and Dialects (VarDial)}, Santa Fe, USA.

\bibitem[\protect\BCAY{Zampieri, Tan, Ljube\v{s}i\'{c},\ \BBA\
  Tiedemann}{Zampieri et~al.}{2014}]{zampieri6}
Zampieri, M., Tan, L., Ljube\v{s}i\'{c}, N., \BBA\ Tiedemann, J.
  \BBOP2014\BBCP.
\newblock \BBOQ {A Report on the DSL Shared Task 2014}\BBCQ\
\newblock In {\Bem Proceedings of the First Workshop on Applying NLP Tools to
  Similar Languages, Varieties and Dialects}, \BPGS\ 58--67, Dublin, Ireland.

\bibitem[\protect\BCAY{Zampieri, Tan, Ljubeši\'{c}, Tiedemann,\ \BBA\
  Nakov}{Zampieri et~al.}{2015}]{zampieri8}
Zampieri, M., Tan, L., Ljubeši\'{c}, N., Tiedemann, J., \BBA\ Nakov, P.
  \BBOP2015\BBCP.
\newblock \BBOQ {Overview of the DSL Shared Task 2015}\BBCQ\
\newblock In {\Bem Proceedings of the Joint Workshop on Language Technology for
  Closely Related Languages, Varieties and Dialects (LT4VarDial)}, \BPGS\ 1--9,
  Hissar, Bulgaria.

\bibitem[\protect\BCAY{Zhong, Chen, Zhu,\ \BBA\ Liu}{Zhong
  et~al.}{2007}]{zhong1}
Zhong, S., Chen, Y., Zhu, C., \BBA\ Liu, J. \BBOP2007\BBCP.
\newblock \BBOQ Confidence measure based incremental adaptation for online
  language identification\BBCQ\
\newblock In {\Bem Proceedings of International Conference on Human-Computer
  Interaction (HCI 2007)}, \BPGS\ 535--543, Beijing, China.

\bibitem[\protect\BCAY{Zlatkova, Kopev, Mitov, Atanasov, Hardalov, Koychev,\
  \BBA\ Nakov}{Zlatkova et~al.}{}]{zlatkova1}
Zlatkova, D., Kopev, D., Mitov, K., Atanasov, A., Hardalov, M., Koychev, I.,
  \BBA\ Nakov, P.
\newblock
\newblock \BBOQ An ensemble-rich multi-aspect approach for robust style change
  detection\BBCQ.

\end{thebibliography}
\bibliographystyle{theapa}

\end{document}